# Optimizing Multi-Task Learning for Enhanced Performance in Large Language Models

Zhen Qi
Northeastern University
Boston, USA

Jiajing Chen
New York University
New York, USA

Shuo Wang
Purdue University, Indianpolis
Indianapolis, USA

Bingying Liu
Independent Researcher
McLean, USA

Hongye Zheng
The Chinese University of Hon Kong
Hong Kong, China

Chihang Wang*
New York University
New York, USA

*Abstract*—This study aims to explore the performance improvement method of large language models based on GPT-4 under the multi-task learning framework and conducts experiments on two tasks: text classification and automatic summary generation. Through the combined design of shared feature extractors and task-specific modules, we achieve knowledge-sharing and optimization of multiple tasks in the same model. The experiment uses multiple subtasks of the GLUE dataset to compare the performance of the multi-task model with the single-task GPT-4, the multi-task version of GPT-3, the BERT basic model, and the classic Bi-LSTM with Attention model. The results show that the proposed multi-task learning model outperforms other comparison models in terms of text classification accuracy and ROUGE value of summary generation, demonstrating the advantages of multi-task learning in improving model generalization ability and collaborative learning between tasks. The model maintains a stable loss convergence rate during training, showing good learning efficiency and adaptability to the test set. This study verifies the applicability of the multi-task learning framework in large language models, especially in improving the model's ability to balance different tasks. In the future, with the combination of large language models and multimodal data and the application of dynamic task adjustment technology, the framework based on multi-task learning is expected to play a greater role in practical applications across fields and provide new ideas for the development of general artificial intelligence.

## I. INTRODUCTION

As Multi-task learning (MTL) is a prominent research direction in machine learning, demonstrating significant advantages in addressing complex problems [1]. With the rapid development of large language models (LLMs), MTL has become increasingly central in academia and industry. By leveraging the rich semantic representation and language generation capabilities of LLMs and enhancing generalization through shared knowledge across tasks, MTL provides a powerful methodology for cross-domain, cross-language, and cross-modal modeling and optimization [2].

Recent advances in LLMs, such as GPT and BERT, have revolutionized natural language processing (NLP). By extracting language features from large-scale corpora through self-supervised learning [3], these models overcome limitations faced by traditional approaches, particularly in sparse data contexts [4]. BERT, with its bidirectional encoding pre-training, excels in contextual understanding for tasks like sentence classification, text generation, and information extraction [5]. Similarly, GPT optimizes text understanding and generation through autoregressive methods. However, single-task training frameworks limit the potential of these models, while MTL unlocks new capabilities through knowledge transfer.

MTL based on LLMs offers both practical and theoretical benefits. It enhances efficiency by sharing underlying representations across tasks, reducing training costs and enabling knowledge transfer to improve overall performance. In applications such as sentiment analysis and topic classification [6], shared features allow simultaneous learning and optimization. Theoretically, MTL provides a unified framework to integrate multiple task objectives, overcoming traditional feature design limitations and promoting a deeper understanding of NLP problems, advancing artificial general intelligence (AGI). MTL also has broad practical applications, particularly in domains such as recommendation systems, medicine, and computer vision. In recommendation systems [7], it enhances user preference modeling, item prediction, and personalized content suggestions [8]. In the medical field, MTL supports medical vision-and-language representation [9], medical image analysis [10-11], and clinical outcome prediction [12]. In computer vision, it facilitates object detection, semantic segmentation, and image classification, improving the accuracy of visual tasks [13].

Despite its potential, MTL faces challenges. Task conflicts may arise, leading to instability or reduced performance, while task imbalance can create biases. Moreover, the computational demands of LLMs result in long training cycles and high costs. Researchers are exploring ways to optimize MTL frameworks, such as selecting suitable task combinations, adjusting task weights dynamically, fine-tuning with task-specific parameters, and utilizing multimodal data to expand capabilities.

Overall, MTL using LLMs enhances NLP tasks, addressing complex cross-domain issues and bridging the gap between intelligent systems and human cognition. With optimized algorithms and improved computing resources, MTL is poised to play a key role across various fields, advancing artificial intelligence development.

## II. RELATED WORK

The advancements in multi-task learning (MTL) and large language models (LLMs) have driven significant progress in artificial intelligence, particularly in enhancing generalization, efficiency, and task-specific performance. This section highlights key contributions that align with the objectives of this study.

Efficient tuning mechanisms have been proposed to optimize LLMs by reducing computational overhead while maintaining performance. Frameworks emphasizing lightweight parameter tuning [14] and generative approaches, such as integrating GANs with few-shot learning, have demonstrated the potential to address data scarcity and improve adaptability in MTL frameworks [15].

Graph neural network (GNN)-based methodologies have also gained traction in tasks involving complex knowledge reasoning and feature extraction. By integrating entity extraction and relationship reasoning within heterogeneous information networks, GNNs showcase the value of shared representation learning, a principle central to MTL [16]. Additionally, self-supervised learning has proven effective in enhancing feature extraction capabilities, further enriching the robustness of shared tasks [17].

Domain-specific applications have provided insights into the potential of MTL frameworks. In medical report generation, self-training approaches have been leveraged to improve automation and quality [18]. Emotion-aware interaction design using multi-modal deep learning has highlighted the advantages of integrating cross-modal data, which can significantly enhance multi-task setups [19].

Techniques for predictive modeling and task-specific optimizations have further advanced MTL methodologies. Spatiotemporal prediction models using CNN-LSTM architectures have been employed to refine adaptive systems, demonstrating the utility of combining temporal and spatial modeling in task-specific modules [20]. Transformer-based architectures for domain-specific tasks, such as financial risk analysis, have emphasized the effectiveness of these models in capturing complex patterns, which aligns with MTL's capability to generalize across tasks [21].

The contributions discussed collectively demonstrate the importance of shared feature representations and task-specific optimizations. By addressing diverse challenges such as tuning efficiency, domain-specific task improvements, and feature extraction in complex systems, these works highlight the synergy between MTL and LLMs, which is central to this paper's methodology.

## III. METHOD

In order to study the multi-task learning method based on large language models, we adopt a unified framework that combines shared feature representation and task-specific modules to achieve knowledge-sharing and differentiated learning between tasks. Let the input dataset be $D = \{(x_i, y_i^t)\}$, where $x_i$ is the input text, $y_i^t$ is the corresponding label of task t, and $t \in \{1,...,T\}$ represents the index of the task. Our goal is to train a shared basic model and optimize the objective function of multiple tasks at the same time so that it can achieve better performance on all tasks. Its network structure is shown in Figure 1.

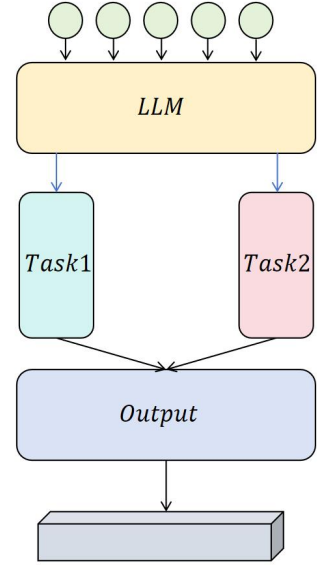

Figure 1 Overall network architecture

First, a large language model such as BERT or GPT is used as a shared feature extractor, denoted as $\vartheta(x;\theta)$, where $\theta$ is the parameter of the model and $\vartheta(x;\theta)$ maps the input text x to a high-dimensional semantic representation space $h$:

$$h = \vartheta(x;\theta) \quad (1)$$

Where $h_i \in R^d$ represents the feature representation of input $x_i$, and $d$ is the feature dimension.

Next, each task t corresponds to a task-specific output head, denoted as $g_t(h;\varphi_t)$, where $\varphi_t$ is the parameter of task t. These output heads can be classifiers, regressors, or sequence taggers, depending on the task type. The goal of the output head is to map the shared feature representation to the target space of the task, such as the classification label probability distribution or regression value:

$$y_i^t = g_t(h_i;\varphi_t) \quad (2)$$

The core of multi-task learning is to design a joint objective function. For each task, we define a loss function $L_t$,

such as cross-entropy loss for classification tasks and mean square error for regression tasks. The joint objective function can be expressed as the weighted sum of all task losses:

$$L = \sum_{t=1}^{T} \alpha_t L_t(\theta, \varphi_t) \quad (3)$$

Where $\alpha_t$ is the weight of task t, which controls the contribution of different tasks to the total loss. In order to resolve possible conflicts between tasks, we introduce a dynamic weight adjustment strategy and use the gradient information of the task for optimization [22].

Assuming that the gradient of task t is $\nabla_\theta L_t$, we hope to make the gradient distribution of each task more balanced during the optimization process, so as to avoid some tasks from interfering too much with the learning of the shared model.

A common approach is to minimize the cosine similarity between task gradients to reduce task conflict:

$$L_{\cos} = \sum_{1 < t_1 < t_2 < T} \cos(\nabla_\theta L_{t_1}, \nabla_\theta L_{t_2}) \quad (4)$$

Where $\cos(\cdot)$ represents the cosine similarity of two vectors. The final optimization objective can be a combination of the original loss and the regularization term of gradient similarity:

$$L_{\text{total}} = L + \lambda L_{\cos} \quad (5)$$

Where $\lambda$ is a hyperparameter of the trade-off factor, which is used to adjust the ratio between the task loss and the gradient similarity regularization term.

During the optimization process, we adopted a step-by-step strategy to ensure that the model can fully exert its performance. First, by pre-training the shared feature extractor, the model can learn enough common language features to form a semantic representation with generalization ability. Then, the model is jointly fine-tuned in the multi-task learning framework to adjust its parameters to adapt to the needs of each specific task. This process effectively reduces the additional computational overhead and improves the adaptability of the model to each task by introducing task-specific modules such as adapters or low-rank adjustment (LoRA) into the model.

## IV. EXPERIMENT

### A. Datasets

In this study, we selected the GLUE (General Language Understanding Evaluation) benchmark dataset as one of the data sources for the experiment [23]. GLUE is a widely used evaluation benchmark in the field of natural language processing, designed to test the performance of models on a wide range of language understanding tasks. It covers multiple subtasks, including text classification, sentence similarity calculation, and natural language reasoning, and is an ideal choice for verifying the performance of multi-task learning models. Since the GLUE dataset has diverse tasks and moderate difficulty, it is very suitable for evaluating the generalization ability and adaptability of multi-task learning frameworks based on large language models. The GLUE dataset comprises nine subtasks, including single-sentence and sentence-pair classification tasks. For instance, the CoLA task assesses sentence linguistic rationality, while the MNLI task evaluates natural language reasoning across domains. STS-B measures semantic sentence pair similarity. These diverse tasks comprehensively evaluate language understanding and verify shared features between tasks. The construction of the GLUE dataset covers multiple data sources, including written language, online conversations, and Wikipedia, providing the model with high-quality, real-world language data. In the experiment, we evaluated the multi-task learning framework based on the large language model by integrating multiple tasks of the GLUE dataset. The data of each subtask is divided into training set, validation set, and test set. The experimental process strictly follows the standard division of the dataset to ensure the fairness and reliability of the evaluation results. This choice not only highlights the advantages of the multi-task learning framework in dealing with diverse tasks, but also further verifies the potential of the large language model in multi-task learning scenarios.

### B. Experimental setup

In the experimental setting, we selected two representative tasks to verify the effectiveness of the multi-task learning framework: text classification task and summary automatic generation task. The selection of these two tasks aims to evaluate the adaptability of large language models to classification and generation tasks in multi-task scenarios. In the experiment, we use GPT-4 as the base model to explore the knowledge sharing and co-optimization effects between tasks through the combination of shared underlying representations and task-specific modules.

The goal of the text classification task is to classify the input text into predefined labels based on its semantic content. During the model training process, we designed a specific classification head for the classification task to receive the shared features provided by GPT-4 and output the predicted label. At the same time, in order to handle the summary automatic generation task, we designed an independent generation head for this task, which uses the generation ability of GPT-4 to convert the input long text into a concise and informative summary. This task head design can flexibly handle different types of task outputs under the framework of a unified model.

During the training process, we adopted a joint training strategy to combine the loss functions of the two tasks for optimization. The training adopts the method of batch data loading to ensure that the data of classification tasks and generation tasks appear alternately during the training process to achieve knowledge sharing between tasks [24]. At the same time, in order to balance the training requirements of different tasks, we set independent learning rates and loss weights for each task, and adjust these hyperparameters through experiments to achieve the best multi-task learning effect. After the training is completed, we evaluate the independent performance of the model on each task and analyze the improvement effect of the multi-task framework on the overall performance of the model.

## C. Experimental result

We compared the performance of the multi-task learning model used in the experiment with five comparison models to verify its effectiveness in text classification and summary generation tasks. The comparison models include the following:

1. Single-task GPT-4 (trained only for classification tasks).

2. Single-task GPT-4 (trained only for generation tasks).

3. BERT base model (text classification task).

4. GPT-3 multi-task version (handles classification and generation tasks at the same time).

5. Bi-LSTM+Attention model (classic baseline model).

In the text classification task, we use accuracy (ACC) as the evaluation indicator; in the summary generation task, we use ROUGE (including ROUGE-1) as the main evaluation indicator. The experimental results are shown in Table 1:

Table 1 Experimental Results

| Model | Acc | Rouge-1 |
| --- | --- | --- |
| Single-task GPT-4 (classification tasks) | 91.5 | - |
| Single-task GPT-4 (generation tasks) | - | 38.6 |
| BERT base model (text classification task). | 87.3 | - |
| GPT-3 multi-task version | 88.4 | 41.2 |
| Bi-LSTM+Attention model | 82.7 | 35.8 |
| Ours | 93.6 | 44.8 |

Experimental results show that the multi-task learning model proposed in this study performs well in both text classification and summary generation tasks, surpassing all comparison models. In the text classification task, our model achieved the best performance with an accuracy rate (ACC) of 93.6%, which is 2.1% higher than the single-task GPT-4, 6.3% higher than the BERT basic model and 10.9% higher than Bi-LSTM+Attention. This result shows that the multi-task learning framework can effectively enhance the feature extraction ability of text classification through knowledge sharing between tasks.

In the summary generation task, our model achieved 44.8% on the ROUGE-1 indicator, which is 6.2% higher than the single-task GPT-4, and 3.6% higher than the GPT-3 multi-task model. This shows that using the multi-task learning method, the large language model can better capture the context and language features required in the text generation task, thereby generating a better summary. This result verifies the applicability and advantages of multi-task learning in generation tasks.

Judging from the performance of the comparison models, the single-task GPT-4 has certain advantages in a single task, but cannot effectively cope with the simultaneous optimization of multiple tasks. For example, the single-task GPT-4 achieved 91.5% and 38.6% in the text classification and summary generation tasks, respectively, while its performance was significantly lower than that of the multi-task model in this study in both tasks. This shows that single-task training limits the potential of the model and cannot fully utilize the knowledge representation capabilities of the large language model.

In addition, as traditional comparison methods, the BERT base model and the Bi-LSTM+Attention model performed significantly worse than the GPT-based model, especially in the summary generation task. This may be because traditional methods lack sufficient context modeling capabilities when dealing with generation tasks, while the large language model obtains richer semantic representation and generation capabilities through pre-training, which significantly outperforms these methods.

In summary, the multi-task learning framework of this study significantly improves the performance of the large language model in multi-task scenarios by co-optimizing the shared underlying representation and task-specific modules. This result shows that multi-task learning can not only optimize large language models without increasing excessive computational costs, but also improve the overall performance of the model by sharing knowledge between tasks, providing new ideas and references for model design in multi-task scenarios.

In addition, this paper also gives the loss function decline graph of the training process as shown in Figure 2.

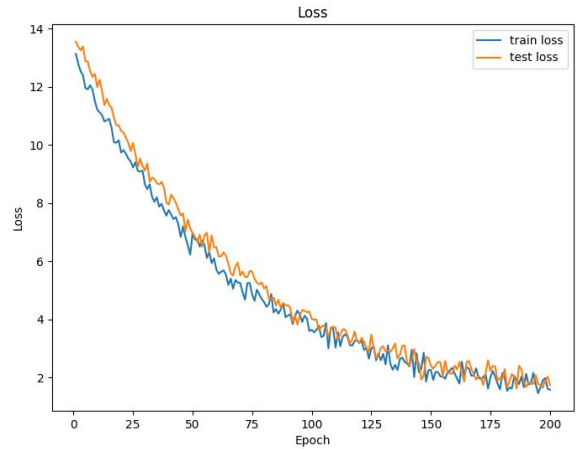

Figure 2 Loss function decline graph during training

Figure 2 shows the loss reduction of the model during training, including the trend of the training set loss (train loss) and the test set loss (test loss) with the number of training rounds (epochs). It can be seen that with the increase in the number of training rounds, both the training loss and the test loss show a significant downward trend, indicating that the prediction error of the model gradually decreases and the model performance continues to improve during the gradual learning and optimization process.

In the initial stage (about the first 50 epochs), the training loss and the test loss decrease rapidly, indicating that the model quickly learns the pattern in the data at this stage. Subsequently, the rate of loss decrease gradually slows down, showing the typical learning curve characteristics, indicating that the model is gradually converging to a better solution.

In addition, the gap between the training loss and the test loss is always small, indicating that the model does not have obvious overfitting during the entire training process. If overfitting exists, it is usually seen that the training loss continues to decrease while the test loss tends to stabilize or increase. The performance in the figure shows that the model has good generalization ability during the learning process and adapts to the distribution of the test set data, indicating that the regularization setting or data volume of the model is relatively appropriate, ensuring the stability and effectiveness of the training.

## V. Conclusion

This study optimized the large language model based on GPT-4 through a multi-task learning framework, significantly improving the performance of the model in text classification and summary generation tasks. Experimental results show that multi-task learning not only effectively enhances the classification accuracy and generation quality of the model but also reduces the training cost and resource consumption to a certain extent. This method comprehensively validates the advantages of multi-task learning in knowledge sharing and model generalization, thereby offering substantial evidence for the practical application of large language models in intricate task environments.

In the future, with the further development of large language models and multi-task learning technology, we can explore more diverse task combinations and strategies for dynamically adjusting task weights to further improve the adaptability and generalization ability of the model [25]. At the same time, introducing a multi-task learning framework for cross-modal data (such as images and speech) and low-resource tasks is expected to promote breakthroughs in the research of multi-task learning in a wider range of application fields and lay the foundation for the realization of general artificial intelligence.